\documentclass{article}

\usepackage{microtype}
\usepackage{graphicx}
\usepackage{subcaption}
\usepackage{booktabs} 
\usepackage{graphicx} 
\usepackage{marvosym}
\usepackage{pdfpages}

\usepackage{xurl}

\usepackage{hyperref}

\usepackage[preprint]{icml2026}

\usepackage{makecell}
\usepackage{amsmath}
\usepackage{amssymb}
\usepackage{mathtools}
\usepackage{amsthm}

\usepackage[capitalize,noabbrev]{cleveref}

\theoremstyle{plain}

\theoremstyle{definition}

\theoremstyle{remark}

\newcommand{\Framework}{MemCoder}

\usepackage[textsize=tiny]{todonotes}
\usepackage{threeparttable}
\usepackage[table]{xcolor}
\usepackage{multirow}
\usepackage{tcolorbox}
\usepackage{listings}
\definecolor{prompt}{RGB}{59, 130, 246}      
\definecolor{iter0}{RGB}{52, 211, 153}       
\definecolor{iter1}{RGB}{251, 191, 36}       
\definecolor{iter2}{RGB}{168, 85, 247}       
\definecolor{iter3}{RGB}{239, 68, 68}        
\usepackage{graphicx}
\usepackage{subcaption} 
\usepackage[T1]{fontenc}
\usepackage{newtxtext}
\usepackage{newtxmath}

\usepackage{tabularx}
\newcolumntype{Y}{>{\centering\arraybackslash}X} 

\icmltitlerunning{Submission and Formatting Instructions for ICML 2026}

\begin{document}

\twocolumn[
  \icmltitle{
 Your Code Agent Can Grow Alongside You with Structured Memory}

  \icmltitlerunning{Your Code Agent Can Grow Alongside You with Structured Memory} 

  \icmlsetsymbol{intern}{*}

  \begin{icmlauthorlist}
    \icmlauthor{Yi-Xuan Deng}{thu,intern}
    \icmlauthor{Xiaoqin Liu}{seer}
    \icmlauthor{Yi Zhang}{thu}
    \icmlauthor{Guo-Wei Yang}{seer}
    \icmlauthor{Shuojin Yang\textsuperscript{\Letter}}{thu}

  \end{icmlauthorlist}

  \icmlaffiliation{thu}{Tsinghua University}
  \icmlaffiliation{seer}{Proxseer Inc} 

  \icmlcorrespondingauthor{Shuojin Yang}{yangshuojin@mail.tsinghua.edu.cn}

  \icmlkeywords{Machine Learning, ICML}

  \vskip 0.3in
]

\printAffiliationsAndNotice{\textsuperscript{*}Intern at Proxseer Inc}

\begin{abstract}

While "Intent-oriented programming" (or "Vibe Coding") redefines software engineering, existing code agents remain tethered to static code snapshots. Consequently, they struggle to model the critical information embedded in the temporal evolution of projects, failing to leverage the "reasoning trajectories" implicit in past successful practices. This limitation results in rigid behavioral logic and a lack of autonomous adaptability, ultimately hindering their ability to tackle complex, repository-level problems. To bridge this static–dynamic mismatch, we propose \Framework{}, a framework designed to enable continual human-AI co-evolution. \Framework{} first structures historical human experience to distill latent intent-to-code mappings from past commits. It then employs a self-refinement mechanism driven by verification feedback to correct agent behavior in real-time. Crucially, an experience self-internalization mechanism is introduced to crystallize human-validated solutions into long-term knowledge, thereby supporting sustained evolution. Experimental results on SWE-bench Verified demonstrate that \Framework{} not only achieves State-of-the-Art (SOTA) performance but also delivers a 9.4\% improvement in resolved rate over the general foundation model DeepSeek-V3.2. These findings indicate that equipping agents with the capability to co-evolve with humans via project history and real-time feedback effectively unlocks the potential of general models in complex software engineering tasks.
\end{abstract}

\section{Introduction}
\label{sec:intro}

The rapid evolution of Large Language Models (LLMs)~\cite{DBLP:journals/corr/abs-2505-09388, guo2025deepseek, geminifamily,zhang2025bee} has fundamentally transformed software engineering into a human-AI collaborative paradigm~\cite{deepseekcoder,hui2024qwen2,glm45}. In this symbiosis, the developer transitions from a manual implementer to a high-level architect, engaging in ``Intent-oriented Programming'' (often referred to as ``Vibe Coding'')~\cite{ge2025survey,DBLP:journals/corr/abs-2511-18538,wang2024survey,cursor, Codegeex, claudecode, wang2024openhands}. This shift positions the human as the source of strategic guidance and constraints, while the agent acts as the dynamic executor. While effective for isolated tasks, this collaboration often fractures in complex, repository-level environments. Here, natural language instructions alone are insufficient to convey tacit knowledge, such as complex inter-file dependencies and unwritten project conventions, that developers have accumulated over time~\cite{pan2025catcoder,repocoder}.

\begin{figure}[t]
  \centering
  \includegraphics[width=\columnwidth]{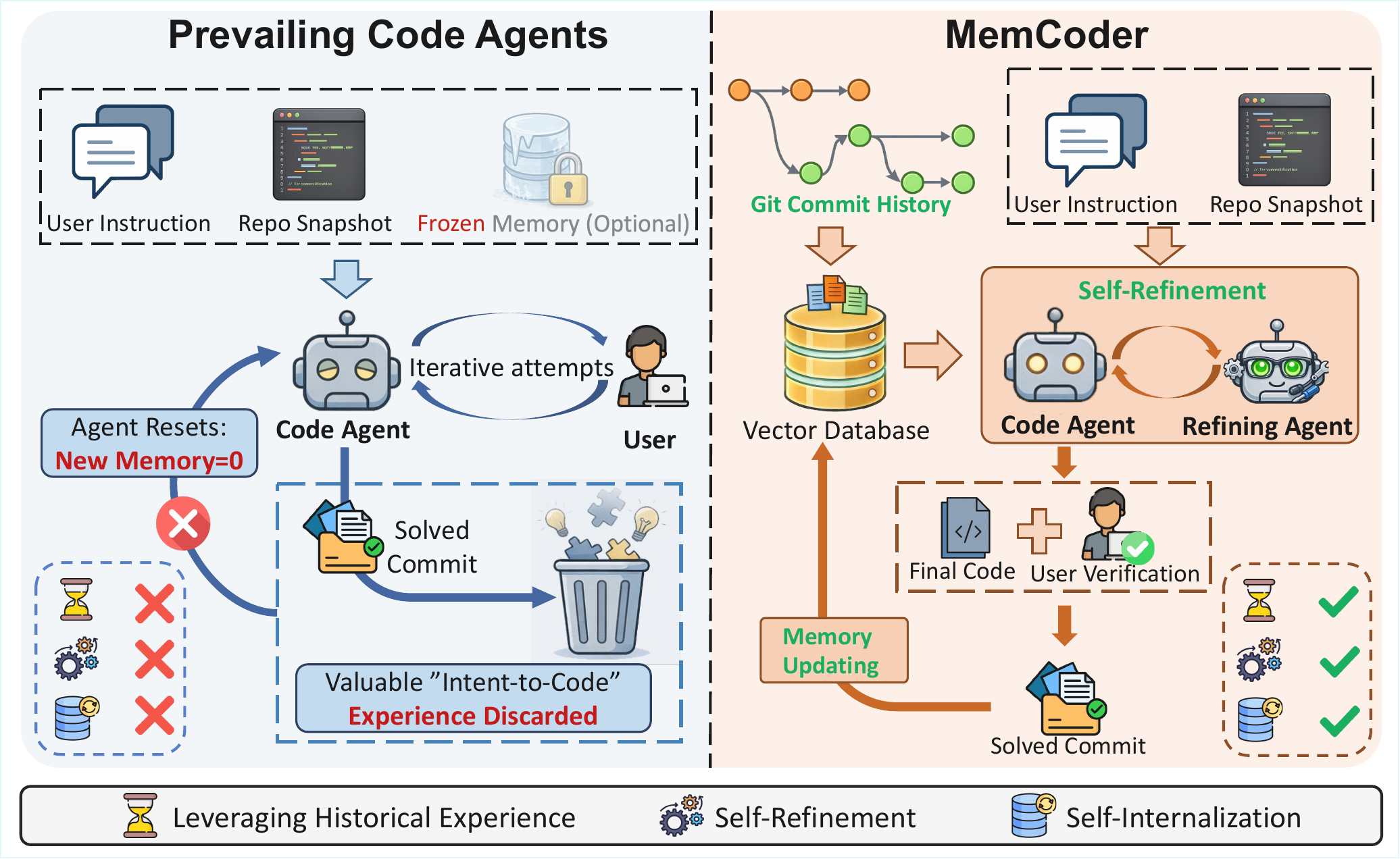} 
  \caption{Comparison of \Framework{} with existing methods. \Framework{} facilitates evolution by learning the intrinsic mapping from high-level intent to concrete code implementation, derived from structured memory.}
  \label{fig:motivation}
\end{figure}

Crucially, these implicit constraints are embedded within the iterative interactions between developers and the codebase. However, as shown in \cref{fig:motivation}, the prevailing code agents~\cite{dong2025survey,wang2024survey,DBLP:conf/ijcai/GuoCWCPCW024} operate under static paradigms that sever the evolutionary feedback loop between human developers and agent capabilities. This disconnection manifests in three critical deficiencies. 
First, current agents overlook the ``human-in-the-loop'' defect-repair patterns archived in version control systems, thereby losing access to historical resolutions of similar conflicts~\cite{cast,RepoHyper,livesweagent}. 
Second, instruction prompting mechanisms are often rigid, failing to bridge the gap between abstract intent and concrete execution. Without retrieving relevant precedents to elaborate on an instruction, these agents fail to inject the implicit details required to align with evolving project standards~\cite{flowgen,CodePori,islam2024mapcoder}. 
Third, and most critically, existing systems fail to internalize human-verified solutions. Consequently, valuable human interventions are discarded rather than integrated, trapping the collaboration in amnesic cycles where the agent repeats errors, forcing the developer to act as a perpetual corrector rather than a co-evolutionary partner~\cite{yao2022react,SWESearch}.

Addressing these limitations necessitates a paradigm shift toward \textit{Human-AI Co-Evolution}. To handle repository-level complexities, an agent must transform from a static executor into an adaptive partner by systematically internalizing human wisdom. This requires a dual approach: reconstructing ``developer cognition'' from historical trajectories to guide decision-making, and crystallizing ephemeral human interactions into enduring capabilities. Such a framework creates a virtuous evolutionary cycle, ensuring the agent progressively attunes itself to the developer's specific coding philosophy and constraints.

Building on these insights, we introduce \Framework{}, a repository-level code agent framework designed for continuous human-AI co-evolution. Recognizing that historical contributions serve as the crucial carrier of both explicit solutions and implicit intent, we instantiate this collaborative paradigm along two dimensions, as illustrated in \cref{fig:motivation}. 
The \textit{Knowledge Dimension} structures unstructured developer commits into memory entries, enabling the agent to recall how humans previously resolved similar complexities. 
The \textit{Execution Dimension} implements intent concretization and dynamic self-refinement: it leverages retrieved historical context to refine sparse instructions into concrete specifications. Subsequently, it crystallizes human-verified solutions into persistent memory to preserve successful intent-to-code mappings. This design ensures that the agent effectively internalizes past human wisdom while iteratively optimizing its alignment with developer intent.

We conducted a comprehensive evaluation of \Framework{} on the SWE-bench Verified dataset~\cite{DBLP:conf/iclr/JimenezYWYPPN24}. The results demonstrate that by empowering agents to co-evolve alongside human developers, \Framework{} achieves State-of-the-Art (SOTA) performance. With GPT-5.2~\cite{openai2025gpt52} as the backbone, our approach sets a new benchmark. Notably, the framework significantly empowers general models like DeepSeek-V3.2~\cite{liu2025deepseek} to better comprehend and replicate human coding patterns, boosting the resolved rate from 68.4\% to 77.8\%.

In summary, the core contributions of this paper are as follows:
\begin{itemize}
    \item We identify the ``interaction disconnect'' in repository-level agents and propose \Framework{}, a framework designed for Human-AI Co-Evolution that enables the agent to grow with the project.
    \item We introduce mechanisms for \textit{Human Experience Internalization}: structuring historical human commits to recall past wisdom, and crystallizing human-verified solutions to accumulate current knowledge.
    \item We achieve SOTA performance on SWE-bench Verified, demonstrating that coupling general models with human-derived evolutionary context significantly empowers them to master complex software engineering tasks.
\end{itemize}

\section{Related Work}

\subsection{LLM-Based Code Generation Agents}
LLM-based code generation agents leverage Large Language Models (LLMs) as their core controllers, achieving autonomous code synthesis through real-time interaction with external tools and environments~\cite{wang2024survey, dong2025survey, DBLP:conf/ijcai/GuoCWCPCW024}. 
Prior research has largely focused on augmenting agent capabilities through specialized workflow designs. 
Inspired by human programming practices, one stream of work adopts iterative optimization driven by generative feedback, as exemplified by Self-Refine~\cite{madaan2023self} and Self-Edit~\cite{zhang2023self}. 
Another stream emphasizes multi-agent collaboration, where role specialization and coordinated execution significantly improve problem-solving efficiency in complex tasks (e.g., AgileCoder~\cite{nguyen2025agilecoder}, AgentCoder~\cite{huang2023agentcoder}, and MapCoder~\cite{islam2024mapcoder}). 
To transcend the limitations of rigid workflows, ReAct~\cite{yao2022react} introduced a paradigm that tightly couples reasoning traces with task-driven actions, serving as a foundational approach for dynamic decision-making.
Building upon this foundational reasoning mechanism, \Framework{} extends the agent's capabilities by incorporating historical development wisdom into its decision-making process, thereby enabling more grounded and effective code synthesis.

\subsection{Long-Term Memory Mechanisms for LLMs}
Long-term Memory (LTM) mechanisms are widely adopted in LLM-based agents to enhance reasoning continuity and multi-agent collaboration~\cite{shan2025cognitive, qian2024experiential}. 
However, continuous interaction leads to the unbounded growth of LTM, posing critical challenges for organization and retrieval. 
A standard strategy decouples LTM from the model, treating it as an external database to facilitate inference~\cite{jiang2024long}. 
To structure this vast information, biologically inspired approaches such as RAPTOR~\cite{sarthi2024raptor} and Memwalker~\cite{chen2023walking} utilize tree-based architectures to organize memory spaces hierarchically. 
Furthermore, methods like MemoryBank~\cite{zhong2024memorybank} and SAGE~\cite{liang2025sage} incorporate the Ebbinghaus forgetting curve into LTM management to maintain conciseness by pruning obsolete information. 
In contrast to these rigid or decay-based structures, \Framework{} introduces a flat, autonomous memory management paradigm that dynamically integrates new insights.

\subsection{Dynamic Evolution}
Dynamic evolution mechanisms enable agents to calibrate their strategies based on historical context and real-time observations. 
This evolution primarily manifests through two pathways: memory consolidation and prompt optimization~\cite{gao2025survey}. 
Specifically, frameworks such as A-mem~\cite{xu2025mem} and MemInsight~\cite{salama2025meminsight} continuously update memory architectures by distilling successful experiences, allowing the agent to achieve self-evolution through iterative environmental interaction. 
In parallel, automated prompt engineering approaches like APE~\cite{zhou2022large} leverage LLMs to generate and validate candidate prompts. 
Subsequent works, including ERM~\cite{yan2025efficient} and OPRO~\cite{yang2023large}, further advance this by introducing optimization driven by execution feedback. 
Distinguishing itself from isolated improvements, \Framework{} integrates diverse evolutionary mechanisms driven by historical human experience to realize a continuous human–AI co-evolution framework.

\begin{figure*}[t] 
    \centering
    \includegraphics[width=\textwidth]{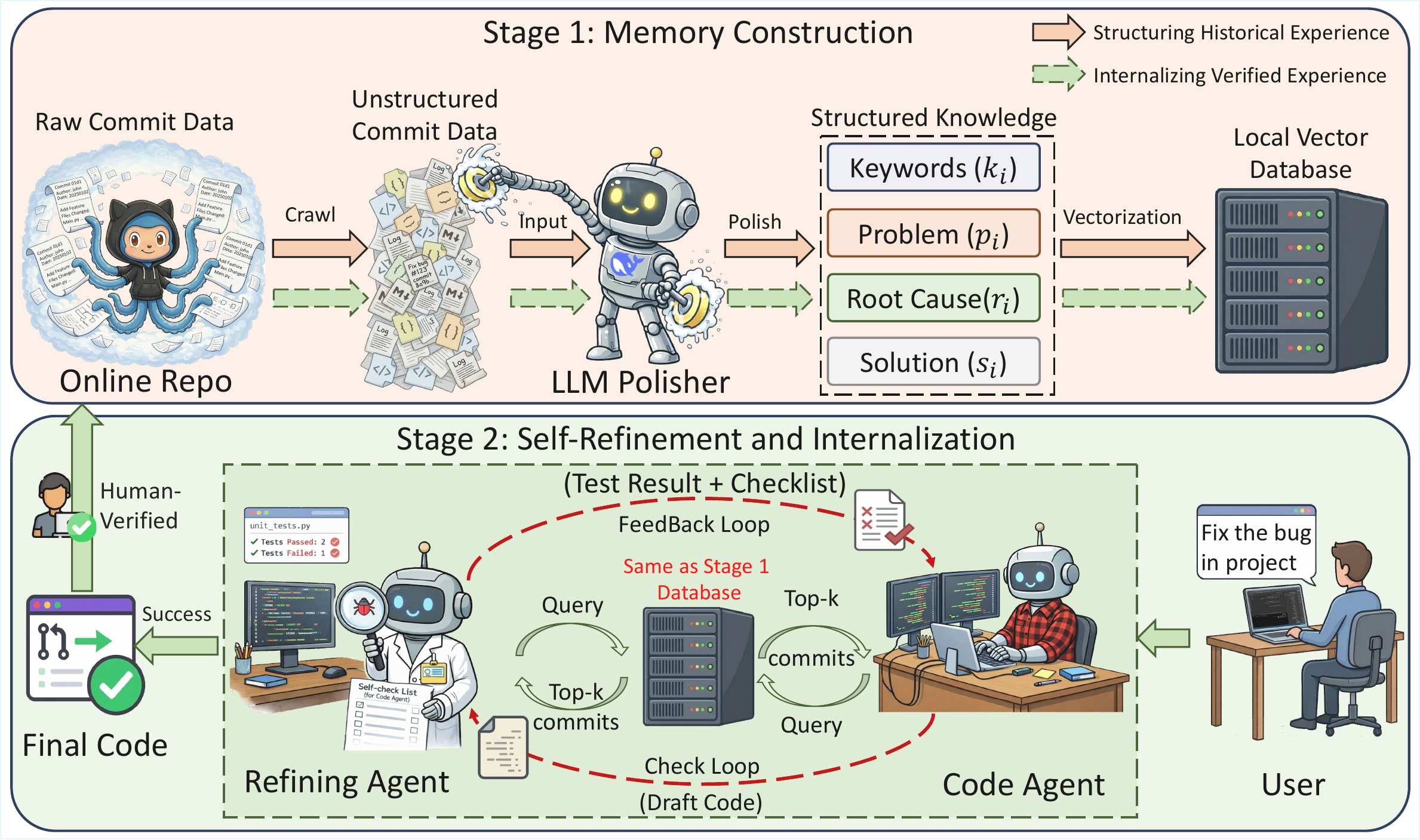} 
    \caption{Architectural overview of \Framework{}, illustrating a closed-loop human--AI co-evolution paradigm.In Stage~1, \Framework{} reconstructs developer cognition by distilling raw commit histories into structured long-term memory, capturing latent intent-to-code mappings from historical human practices. In Stage~2, the agent performs context-aware dual-stage retrieval to access relevant experience, while a Refining Sub-agent enables execution-time self-refinement through prompt concretization, automated test generation, and verification feedback. Crucially, human-validated solutions are subsequently internalized into long-term memory, closing the evolutionary loop and enabling the agent to progressively align with repository-specific conventions across iterations.}
    \label{fig:pipeline}
\end{figure*}

\section{Method}
The core methodology of \Framework{} is built upon a human-AI co-evolution paradigm. Rather than treating code generation as a static inference task, we model it as a continuous learning process where the agent system $\Pi$ iteratively refines itself. This refinement is governed by an evolution function $\text{Evolv}$, which synthesizes insights from long-term memory $M$, current execution trajectories $\tau$, and multi-source feedback $\xi$:

\begin{equation}
\Pi' = \text{Evolv}(\Pi, \mathcal{F}(M, \tau, \xi)).
\label{eq:evolution}
\end{equation}

Here, $\mathcal{F} = \{ f_{\text{struct}}, f_{\text{refine}}, f_{\text{intern}} \}$ represents the composite adaptive mechanism, where the specific formulation of each function is detailed in the following subsections.
To implement this theoretical model, \Framework{} orchestrates three critical phases: constructing structured memory from historical repositories, retrieving context-aware experience during execution, and internalizing human-validated solutions for future iterations.

\subsection{Experience Representation and Utilization}

To bridge the gap between abstract intents and concrete project conventions, we leverage the codebase's history. \Framework{} replaces the fixed snapshot view with a dynamic repository, where raw commits are distilled into reference solutions. This allows the agent to retrieve and replicate the reasoning behind past successful implementations precisely.

\paragraph{Structuring Historical Experience.} High-quality long-term memory serves as the fundamental cornerstone for the human-AI co-evolution of agent systems. However, raw human experiences are often permeated with non-standardized, ambiguous, and idiosyncratic descriptions, which introduce inherent biases during the agent's comprehension and knowledge acquisition. To transform these raw experiences into standardized and agent-centric memory representations for more effective learning, we propose an LLM-driven memory construction algorithm predicated on Defect Management theory~\cite{SETSE052013005}. Specifically, we formalize the memory set as $M=\{m_1, m_2, \dots, m_N\}$, where each memory entry $m_i$ is defined as a structured sextuple:

\begin{equation}
m_i = (o_i, c_i, k_i, p_i, r_i, s_i).
\end{equation}

As shown in Stage~1 of \cref{fig:pipeline}, \Framework{} first ingests the raw historical repository data, including the original commit message $o_i$, the associated code changes $c_i$. Building upon this context, the LLM-driven engine reconstructs the experience by synthesizing a semantic knowledge layer to facilitate agent comprehension. This layer comprises four high-level components designed to transform implicit developer expertise into explicit, agent-friendly knowledge. Specifically, a set of functional keywords $k_i$ abstracts the core phenomena and application scenarios to serve as semantic anchors for precise indexing, while the formal description of the addressed problem $p_i$ captures observable symptoms and situational constraints to further refine the retrieval granularity. 

Building upon these indexing elements, the root-cause analysis $r_i$ elucidates the fundamental logic and technical bottlenecks underlying the issue, enabling the agent system to internalize critical development expertise and reason about the "why" of the problem. Finally, the summarized solution $s_i$ distills the corrective actions into standardized, actionable guidelines, effectively instructing the agent on optimal decision-making and execution procedures when navigating analogous scenarios in future tasks.

\begin{equation}
k_i, p_i, r_i, s_i \leftarrow \text{LLM}(o_i, c_i \mid \text{P}_{gen}).
\end{equation}

To operationalize this construction, we leverage the reasoning capabilities of an LLM to distill latent development intents and issue-fixing expertise embedded within the code changes. The transformation from raw context to the structured sextuple is executed through a task-specific prompt template $\text{P}_{gen}$, which ensures that the resulting memory units are both standardized and optimized for robust agentic reasoning and actionable insight extraction.

Consequently, the memory construction function $f_{struct}$ is defined as:


\begin{equation}
\label{eq:memory_construct}
\begin{split}
    g_i &= \text{LLM}(o_i , c_i \mid \text{P}_{\text{gen}}) \\
    M &\leftarrow \left\{ (o_i,c_i) \oplus g_i \mid (o_i, c_i) \in \mathcal{H} \right\},
\end{split}
\end{equation}

which indicates that the agent transforms the collection of historical commits $\mathcal{H}$ into high-quality long-term memory through LLM and a prompt template.

\paragraph{Context-Aware Dual-Stage Retrieval.} Long-term memory serves as a vast knowledge repository for autonomous agent systems, and a precise retrieval mechanism acts as the critical interface for accessing this information. The quality of content generated by agents, including code synthesis, test case generation, and prompt augmentation, is fundamentally constrained by the relevance of the retrieved memories. Consequently, developing an effective memory retrieval mechanism is of paramount importance.
To ensure scalable and efficient retrieval from extensive long-term memory, we employ an embedding model to encode each reconstructed memory entry $m_i$ into a dense, high-dimensional vector. This process constructs a vectorized database $E = \{\mathbf{e_1}, \mathbf{e_2}, \dots, \mathbf{e_N}\}$, defined as:

\begin{equation}
    \mathbf{e_i} = \operatorname{Embed}(k_i \oplus p_i).
\end{equation}

We implement a two-stage "retrieval-then-rerank" pipeline to identify historical memories most relevant to the current problem. In the first stage, given a search query $q$ derived from the problem description, we perform a rapid approximate nearest neighbor (ANN) search~\cite{aumuller2020ann}, implemented using Facebook AI Similarity Search (FAISS)~\cite{douze2025faiss}. This stage retrieves a candidate set $\mathcal{I}$ comprising a selection of the most similar entries based on cosine similarity.

\begin{equation}
    \mathcal{I} = \operatorname*{arg\,top}_{i \in \{1, \dots, N\}} \left( \frac{\mathbf{q}^\top \mathbf{e}_i}{\|\mathbf{q}\| \|\mathbf{e}_i\|} \right).
\end{equation}

It should be noted that $\mathcal{I}$ encompasses a broader pool of entries than the final selection. By maintaining a robust candidate set, the system effectively mitigates the risk of recall loss during the initial retrieval phase, thereby safeguarding the quality of the filtered historical experiences.

In the second stage, to address the semantic bottleneck inherent in bi-encoder architectures, we employ a cross-encoder reranker for fine-grained semantic matching. This model processes the raw text of both the query and the candidate entries to capture nuanced semantic dependencies. Unlike the initial retrieval phase, the final ranking is determined exclusively by the reranking score:

\begin{equation}
    \alpha_i = \operatorname{CrossEnc}(q,k_i \oplus p_i) \quad \forall i \in \mathcal{I}.
\end{equation}

The search query $q$ is autonomously synthesized by the agent based on its internal reasoning state and task context. This mechanism enhances the precision and adaptivity of the retrieval process, empowering the agent to proactively explore historical knowledge through nuanced formulations tailored to the specific requirements of the issue at hand.

By utilizing the reranker's output as the primary metric for relevance, the system effectively filters out false positives from the initial ANN search, providing the agent with high-fidelity historical insights to guide the subsequent code generation process.

\subsection{Self-Refinement and Internalization} 
The capacity for continual learning from itself and humans is widely recognized as a quintessential hallmark of agentic evolution. Constrained by the inherent staticity of backbone LLMs, agent systems must explore alternative dimensions to facilitate evolution. To this end, \Framework{} investigates this potential through the dual perspectives of prompt optimization and memory updating. The technical details are elaborated as follows.

\paragraph{Dynamic Self-Refinement.} During execution, long-term memory is typically treated as immutable, limiting an agent’s ability to adapt to real-time feedback and to interpret repository-specific conventions.
To address this limitation, \Framework{} introduces a Refining Sub-agent, proactively invoked by the Primary Agent (Stage~2 in \cref{fig:pipeline}), which retrieves human-validated successful experiences from structured memory to ground execution in repository-specific patterns and proven intent-to-code mappings.

Leveraging the retrieved context, the Refining Sub-agent synthesizes targeted test code $t$ and iteratively constructs a verification checklist $l$. By aligning current execution with historically successful practices, this process concretizes developer intent beyond abstract instructions. The resulting checklist directly guides the subsequent decision-making of the Primary Agent, enabling execution-time refinement while remaining consistent with the repository’s evolutionary trajectory.

Concretely, the Refining Sub-agent employs a specialized prompt template $\text{P}_{refine}$ to guide the LLM in generating both the test code $t$ and the verification checklist $l$. The generation is conditioned on a multidimensional context integrating the problem description $p$, execution trace $\tau$, environmental feedback $\xi$, and retrieved historical experiences from memory $M$, formalized as:
\begin{equation}
t, l \leftarrow \text{LLM}(p,\tau,\xi \mid M,\text{P}_{refine}).
\end{equation}
Through iterative invocation, the Refining Sub-agent updates its verification logic by reconciling execution feedback with retrieved successful precedents, allowing the agent to deepen its understanding of the codebase and generate higher-quality, repository-aligned code without modifying long-term memory during execution.
During this phase, the adaptive refinement function $f_{refine}$ updates the feedback state:

\begin{equation}
\label{eq:self-refine}
\xi' \leftarrow \xi \cup \{t, l\},
\end{equation}

highlighting the Refining Sub-agent as the primary driver of execution-time adaptation and a key enabler of human–AI co-evolution within \Framework{}.

\paragraph{Self-Internalization for Experience.} \Framework{} updates its long-term memory by integrating human-verified experiences, thereby preserving the critical alignment between high-level intent and concrete code implementation. At this stage, the memory internalization function $f_{intern}$ is formulated as:

\begin{equation}
\label{eq:memory_update}
    M_{N+1} \leftarrow M_{N} \cup \{m_{N+1}\},
\end{equation}

signifying the realization of a closed-loop evolution process within the agent system. Unlike historical commits, these experiences originate from model-generated solutions and human validation, enabling the memory to gradually shift from human-only priors to human–agent co-evolved knowledge.

\section{Experiment}

\begin{figure*}[t] 
  \centering
  \includegraphics[width=\textwidth]{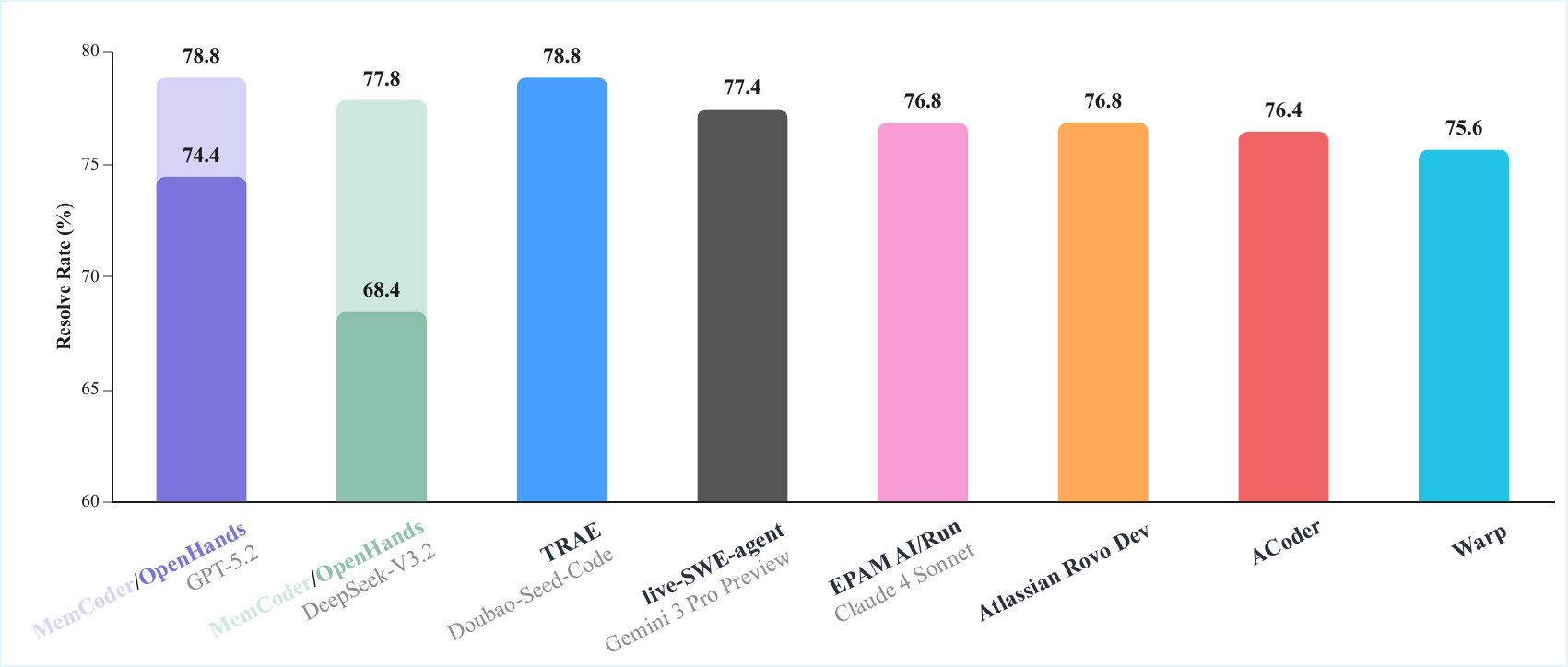} 
  \caption{Comparison of \Framework{} with the top 6 methods on the SWE-bench Verified leaderboard as of January 20, 2026.}
  \label{fig:swe_bench_verified}
\end{figure*}

In this section, we conduct extensive experiments to demonstrate the effectiveness of our proposed method and evaluate the contribution of each individual component to the overall performance of the agent system.

\subsection{Experimental Settings}

\paragraph{Benchmarks.} To evaluate the effectiveness of the proposed framework on real-world software engineering tasks, we adopt SWE-bench~\cite{DBLP:conf/iclr/JimenezYWYPPN24} as our primary benchmark. SWE-bench is a large-scale, execution-based benchmark designed to assess the end-to-end software engineering capabilities of LLM-based agents. It consists of 2,294 task instances collected from real GitHub issues and corresponding pull requests across 12 widely used open-source Python repositories. The benchmark poses significant repository-level challenges, requiring agents to reason over entire codebases, handle complex cross-file dependencies, and generate functional code patches to resolve bugs or implement features. Evaluation is performed by executing generated patches against unit test suites, emphasizing functional correctness rather than textual similarity, and thus providing a rigorous measure of agents’ autonomous problem-solving ability in practical development settings.

To improve evaluation efficiency, we conduct our experiments on SWE-bench Verified, a manually curated, high-quality subset of 500 task instances widely used by existing baselines. Compared to the full benchmark, SWE-bench Verified offers more precise problem specifications and more rigorous test designs, enabling a more reliable assessment of the code patch generation quality of agent frameworks.

\paragraph{Models.} Our experiments are primarily conducted using DeepSeek-V3.2~\cite{liu2025deepseek} as the backbone LLM, selected for its exceptional code generation and tool-calling capabilities. It serves as an ideal choice for the core reasoning engine within an LLM-based autonomous programming framework. Additionally, we verify the absolute performance of our method using the more powerful GPT-5.2 ~\cite{openai2025gpt52}. Experimental results demonstrate that our approach achieves performance levels comparable to the current State-of-the-Art (SOTA) on the SWE-bench Verified leaderboard.

\paragraph{Metrics and Baselines.} We follow the official evaluation methodology provided by SWE-bench. Task instances are executed locally to generate code patches in git diff format, which are then submitted via the official \texttt{sb-cli} tool for evaluation. We report the primary metrics: the resolution rate and the number of resolved issues.  For the SWE-bench Verified dataset, we benchmark our method against the top six performing approaches on the official leaderboard, ensuring a comprehensive comparison with the current state of the field. To ensure a fair evaluation and prevent temporal leakage, we strictly restrict the agent at test time to retrieve only historical experiences created prior to the corresponding test issue.

\subsection{Performance on Repository-Level Code Generation}

On the SWE-bench Verified benchmark, \Framework{} demonstrates robust software engineering code generation capabilities, achieving performance on par with the SOTA. In \cref{fig:swe_bench_verified}, we report the performance of  \Framework{} on GPT-5.2 and provide a comparison with the top six methods~\cite{wang2024openhands,openai2025gpt52,liu2025deepseek,trae,livesweagent,seedcoder,gemini3pro,anthropic2025claude4} on the current leaderboard and backbone models. We conduct experiments under the pass@1 evaluation protocol. The experimental results show that \Framework{} delivers exceptional performance, reaching a level comparable to existing SOTA approaches.

\Framework{} is built upon the OpenHands framework~\cite{wang2024openhands}. To provide a clearer illustration of the performance gains introduced by our approach, \cref{tab:openhands} summarizes the performance of various LLMs~\cite{liu2025deepseek,openai2025gpt52,anthropic2025claudesonnet45,anthropic2025claudepous45,gemini3pro} integrated with the OpenHands agent framework and \Framework{} on the SWE-bench Verified benchmark, with all data sourced from the official OpenHands. While the official OpenHands results are reported under pass@$3$, \Framework{} achieves a solved rate of \textbf{83.8\%} with GPT-5.2 under the more restrictive pass@$2$ setting, highlighting improved efficiency per attempt.

\begin{table}[t]
  \centering
  \renewcommand{\tabcolsep}{1.2mm}
  \caption{Comparative analysis of LLM performance under OpenHands vs. \Framework{} on SWE-bench Verified.}
  \label{tab:openhands}
  \begin{tabular}{lcc}
    \toprule
    Method & Setting & Resolved(\%) \\
    \midrule
    \Framework{} + GPT-5.2 & pass@$2$ &83.8 (419)  \\
    \Framework{} + GPT-5.2 & pass@$1$ &78.8 (394)  \\
    \Framework{} + DeepSeek-V3.2 & pass@$1$ & 77.8 (389)  \\
    OpenHands + Claude Opus 4.5 & pass@$3$ & 77.6 (388)  \\
    OpenHands + Claude Sonnet 4.5 & pass@$3$ & 74.6 (373)  \\
    OpenHands + GPT-5.2 & pass@$3$ & 74.4 (372)  \\
    OpenHands + Gemini 3 pro & pass@$3$ & 70.4 (352)  \\
   
    \bottomrule
  \end{tabular}
\end{table}

\subsection{Ablation Study}
To investigate the individual contributions of various modules within \Framework{}, we conducted a series of ablation studies. Specifically, we evaluated several variant agent configurations by systematically removing the commit retrieval module, the experience representation module, and the dynamic self-refine module. These experiments were performed using DeepSeek-V3.2 as the backbone model on the SWE-bench Verified benchmark. The experimental results are summarized in \cref{tab:main_ablation}.

\begin{table}[h]
  \centering
  \caption{Ablation study of \Framework{} on the SWE-bench Verified dataset using DeepSeek-V3.2. The notation ‘w/o’ indicates experiments where specific modules were removed. The abbreviations CR, ER, and DSR denote the commit retrieval module,  experience representation module, and the dynamic self-refine module, respectively.}
   \label{tab:main_ablation}
  \begin{tabular*}{\columnwidth}{@{\extracolsep{\fill}}lcc}
    \toprule
    Method & Resolved(\%) & $\Delta$ \\
    \midrule
    \Framework{}   & 77.8 (389)  & - \\
    w/o DSR        & 76.4 (382)  & -1.4\%(-7) \\
    w/o DSR \& ER  & 73.0 (365)  & -4.8\%(-24) \\
    w/o CR         & 71.6 (358)  & -6.2\%(-31) \\
    w/o all        & 68.4 (342)  & -9.4\%(-47) \\
    \bottomrule
  \end{tabular*}
\end{table}

\cref{tab:main_ablation} demonstrates that all three proposed modules contribute positively to the performance enhancement of \Framework{}. Notably, the retrieval module yields the most significant performance gains for the overall framework, underscoring that the self-evolving capability of agentic systems can effectively enhance their efficacy in repository-level code generation. To provide deeper insights into the underlying mechanisms of the retrieval module, we perform a comprehensive exploration across three key dimensions: experience representation, retrieval granularity, and retrieval quantity.

\paragraph{Experience Representation.} \Framework{} introduces an LLM-based experience construction mechanism to standardize and structure raw commits within code repositories, ensuring that the vast scale of historical data remains "agent-friendly." To evaluate the actual performance gains derived from structured experiences, we conducted a controlled comparison across three distinct memory configurations while keeping other variables constant: (i) No historical experience; (ii) Raw commits and patch records without structuring; and (iii) Structured memory polished by an LLM. As shown in \cref{tab:main_ablation}, the resolved rate for the "w/o all," "w/o DSR \& ER," and "w/o DSR" settings are 68.4\%, 73\%, and 76.4\%, respectively. These results indicate that even raw, unstructured historical data can provide a baseline performance boost compared to the zero-experience setting. However, the efficiency of information extraction is often hampered by noise, semantic ambiguity, and the inherent gap between human linguistic habits and the processing patterns of backbone LLMs. In contrast, structured memory with a unified style yields more significant and stable performance improvements. These empirical findings substantiate that structured memory is pivotal for enhancing the overall performance of \Framework{}.

\begin{figure}[h] 
  \centering
 
  \includegraphics[width=\columnwidth]{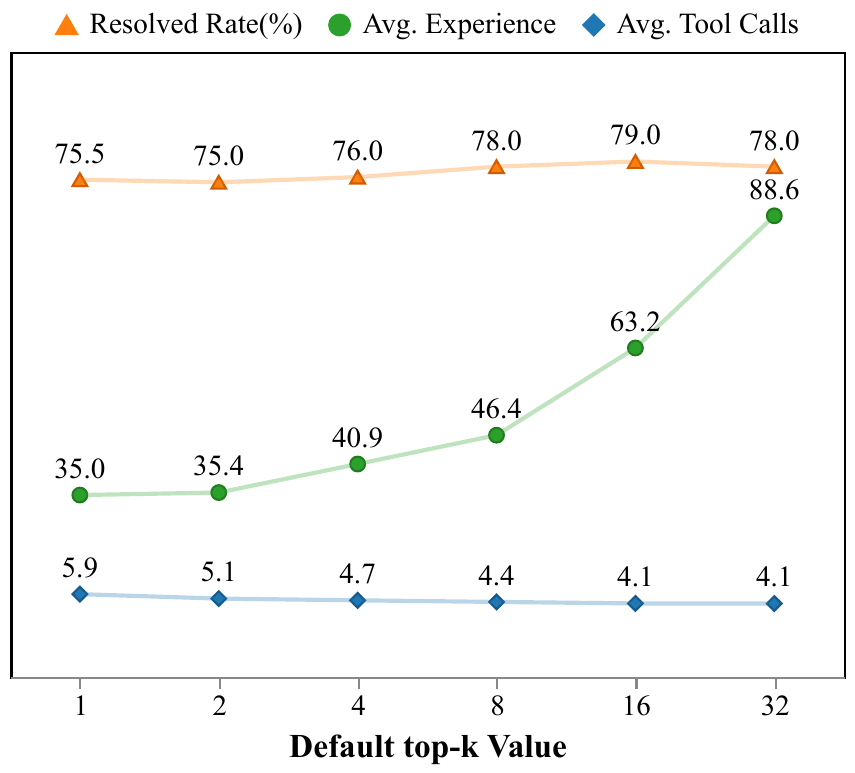} 
  \caption{Performance of \Framework{} across various \texttt{top-k} values. The metrics include resolved rate, average number of retrieved historical experiences, and average tool call frequency. All experiments are conducted using DeepSeek-V3.2 as the backbone model and evaluated on a randomly sampled 200-instance subset of the SWE-bench Verified dataset.}
  \label{fig:ablation1}
\end{figure}

\paragraph{Retrieval Granularity.} While the agent framework autonomously determines the timing of retrieval tool invocation and the scope of information acquisition, we can modulate the agent's behavior by manipulating the initial retrieval parameter, \texttt{top-k}. By adjusting the initial \texttt{top-k}, we investigate the non-linear effects of retrieval granularity on the agent's dynamic interaction patterns and decision-making quality. \cref{fig:ablation1} reveals a significant trade-off between initial information bandwidth and iterative retrieval frequency. 

With a smaller initial \texttt{top-k}, the agent framework compensates by increasing retrieval frequency to ensure sufficient information acquisition. Conversely, as the initial \texttt{top-k} increases, the retrieval frequency gradually declines. This indicates that a broader initial field of view effectively enhances the information "hit rate" per retrieval, enabling the agent to terminate the search process earlier. Concurrently, a marginal increase in retrieved information volume correlates with improved agent performance, suggesting that appropriately augmenting the agent's access to historical experience enhances its capabilities. However, this gain in interaction efficiency does not translate into a sustained linear growth in the resolved rate; the performance curve tends to plateau at larger \texttt{top-k} values. Furthermore, the retrieval frequency ceases to decline, indicating an intrinsic need for the agent to retrieve multi-faceted information to support task completion. Meanwhile, an excessively large \texttt{top-k} significantly dilutes the signal-to-noise ratio of valid evidence, introducing unnecessary context overhead. Thus, the data suggest that a moderate initial retrieval granularity serves as the most effective configuration, maximizing performance while avoiding the diminishing returns and noise associated with excessive information accumulation.

\begin{figure}[h] 
  \centering
 
  \includegraphics[width=\columnwidth]{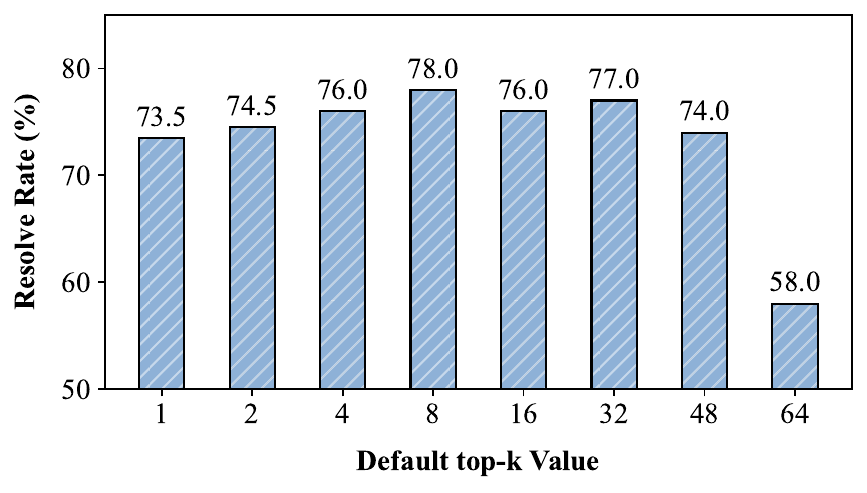} 
  \caption{Performance of \Framework{} across various \texttt{top-k} values with the retrieval frequency restricted to one. The metric displayed is the Resolve Rate (\%). All experiments are conducted using DeepSeek-V3.2 as the backbone model and evaluated on a randomly sampled 200-instance subset of the SWE-bench Verified dataset.}
  \label{fig:ablation2}
\end{figure}

\paragraph{Retrieval Quantity.} To intuitively demonstrate the impact of the quantity of retrieved historical experiences on the performance of the agent system, we constrained the agent to a single retrieval tool invocation and observed performance variations by modulating the \texttt{top-k} parameter (the number of results returned). \cref{fig:ablation2} illustrates that within a lower \texttt{top-k} range, increasing the retrieval scale yields steady performance improvements. This suggests that, at this stage, a greater volume of feedback enables the agent to acquire more historical experiences conducive to resolving the current problem. However, as \texttt{top-k} exceeds 8, the rate of performance gain diminishes, eventually plateauing within a specific range. This implies that with a single retrieval opportunity, the agent has already sufficiently acquired the most representative experiences, rendering additional retrieval results redundant. Conversely, when \texttt{top-k} becomes excessively large, the agent suffers from performance degradation due to an overload of context information. 

These results underscore the "signal-to-noise ratio" challenge inherent in Retrieval-Augmented Generation (RAG): while expanding the retrieval scope enhances the probability of recalling relevant information, it simultaneously dilutes the effective density of the context. An excessively large \texttt{top-k} leads to the accumulation of irrelevant segments, which not only taxes the model's attention mechanism but also exacerbates the "Lost-in-the-Middle" phenomenon, resulting in the degeneration of inference capabilities. Consequently, rather than indiscriminately increasing the volume of retrieval, prioritizing the precision of retrieval results proves more critical for enhancing agent performance.

\section{Conclusion}

We proposed \Framework{} to address the limitation where static code agents fail to capture the critical information embedded in the temporal evolution of projects. Our approach reconstructs developer cognition by structuring historical commits and leveraging this retrospective wisdom to guide the agent in verifying and refining its own execution. Experimental results on SWE-bench Verified demonstrate that \Framework{} achieves State-of-the-Art performance and effectively unlocks the potential of general models like DeepSeek-V3.2 in complex engineering tasks. These findings validate the critical importance of Human-AI Co-Evolution. By continuously internalizing the reasoning trajectories embedded in human history, the agent transcends its role as a mere executor to become an adaptive partner capable of growing alongside the developer.

\section{Impact Statement}
This paper presents work whose goal is to advance the field of machine learning. There are many potential societal consequences of our work, none of which we feel must be specifically highlighted here.
\bibliography{example_paper}
\bibliographystyle{icml2026}

\appendix
\onecolumn
\section*{APPENDIX}

\section{Case Study}

To intuitively illustrate the impact of retrieving historical experience on the agent's behavior, we select the instance with \texttt{instance\_id} \texttt{django\_\_django-16315} as a representative example. As shown, without access to the retrieved experience, the agent applies fixes at an incorrect location. In contrast, the historical experience highlights a prior modification to the function \texttt{on\_conflict\_suffix\_sql}, effectively signaling an issue in the current code’s invocation of this function, which in turn guides the agent to successfully repair the bug.

\begin{center}
\includegraphics[
  page=1,
  height=0.8\textheight,
  keepaspectratio
]{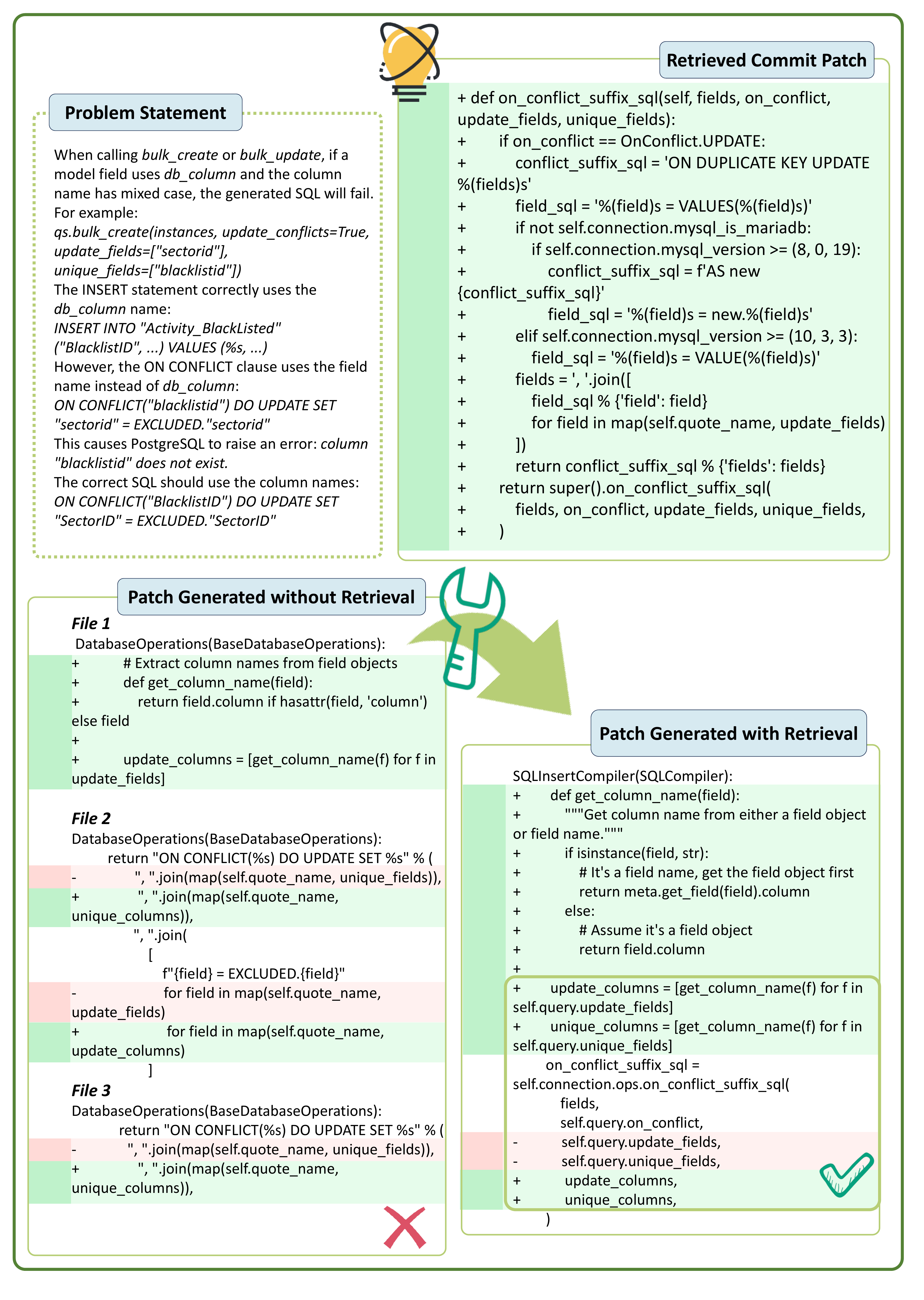}
\end{center}

\section{Algorithmic Pipeline}
\label{alg:algorithmic pipeline}

\cref{alg:algorithmic pipeline} provides the pseudocode of the proposed pipeline as a concise algorithmic summary. The pseudocode formalizes the overall control flow and module interactions of our method, including memory construction, experience retrieval, iterative refinement, and experience internalization with human assistance. All components and design choices follow the pipeline described in the main text, and the pseudocode is intended solely to clarify execution order and interface dependencies, rather than to introduce additional mechanisms.

\begin{algorithm}[H]
\caption{MemCoder Process}
\label{alg:memcoder}
\begin{algorithmic}[1]
    \REQUIRE Target Repo $\mathcal{R}$, Issue $P$, Existing Memory $\mathcal{M}$
    \ENSURE Resolved Solution $S$
    
    \STATE \textbf{Definitions:} 
    \STATE \quad $\mathcal{C}$: Context history (actions, observations, thoughts)
    \STATE \quad $c$: Candidate code patch generated by agent
    
    \item[] 
    \STATE \textbf{Stage 1: Memory Synchronization}
    \STATE Identify new commits $H_{new}$ in $\mathcal{R}$ that are not in $\mathcal{M}$
    \FORALL{$h_i \in H_{new}$}
        \STATE $m_i \leftarrow$ \textsc{LLMPolish}($h_i$) \COMMENT{Summarize commit logic}
        \STATE $v_i \leftarrow$ \textsc{Embed}($m_i$)
        \STATE $\mathcal{M} \leftarrow \mathcal{M} \cup \{ (m_i, v_i) \}$ \COMMENT{Update memory}
    \ENDFOR

    \item[] 
    \STATE \textbf{Stage 2: Execution Loop}
    \STATE $\mathcal{C} \leftarrow \{ P \}$, $solved \leftarrow$ \textbf{false}
    
    \WHILE{not $solved$ \AND steps $<$ MaxSteps}
        \STATE $Action, Args \leftarrow$ \textsc{PrimaryAgent}($\mathcal{C}$)
        
        \IF{$Action ==$ "Retrieve"}
            \STATE $K \leftarrow$ \textsc{Retrieve}($Args, \mathcal{M}$)
            \STATE $\mathcal{C} \leftarrow \mathcal{C} \cup K$
        \ELSIF{$Action ==$ "Refining Agent"}
            \STATE $c \leftarrow Args$ 
            \STATE $TestRes, Checklist \leftarrow$ \textsc{Refining Agent}($c, P, \mathcal{M}$)
            
            \IF{$TestRes$ is \textbf{Pass}}
                \STATE $S \leftarrow c$
                \STATE $solved \leftarrow$ \textbf{true}
                \STATE \textbf{break}
            \ELSE
                \STATE $\mathcal{C} \leftarrow \mathcal{C} \cup \{TestRes, Checklist\}$
            \ENDIF
        \ELSE
            \STATE Execute standard tool ($Action, Args$)
            \STATE Update $\mathcal{C}$ with observation
        \ENDIF
    \ENDWHILE
       \STATE \textbf{return} $S$  
    \item[] 
  \STATE \textbf{Stage 3: Submission (Closed-Loop)}
\IF{$solved$}
    \STATE \textbf{Human Review:} \COMMENT{The solution $S$ undergoes human review before the commit is finalized}
    \STATE \textsc{GitCommit}($\mathcal{R}, S$) 
    \STATE \COMMENT{This new commit will be learned in Stage 1 next run}
\ENDIF

\end{algorithmic}
\end{algorithm}




\section{Prompt of Refining Agent}
\label{prompt of refining agent}
This text serves as an explanatory note for \cref{prompt of refining agent}. The Refining Agent is a core component of \Framework{}, responsible for generating test code and synthesizing signals from test outcomes, environment interaction feedback, and retrieved historical experience to produce a clear and actionable instruction checklist for the primary agent. The quality of the generated tests and instruction checklist directly affects the overall performance of the agent; therefore, we provide the Refining Agent with detailed guidance in its prompt.

\begin{lstlisting}

You are a specialized Code Debugging Analysis Agent, designed to analyze code changes and provide detailed debugging insights.

<ROLE>
Your primary role is to thoroughly analyze code patches, identify potential bugs and edge cases, retrieve relevant historical context, provide specific and actionable debugging recommendations, and clearly document the entire analysis process.
</ROLE>

<INPUT_MODES>
You will typically receive two high-level inputs in your context:
- A natural-language **problem description**.
- A **git patch string** (referred to as `git_patch`). **This parameter is always provided and is non-empty.**

Your job is to treat `git_patch` as the candidate fix for the described problem, and you MUST do BOTH:
1) Generate a minimal, runnable reproduction test (or tests) for the problem and show how to run them.
2) Provide a step-by-step debug / review flow to validate and iterate on the patch.
</INPUT_MODES>

<AVAILABLE_TOOLS>
You have access to the following specialized debugging tools:
**execute_bash**, **str_replace_editor**, **think**, **finish**, **execute_ipython_cell**, **task_tracker** and **repo_commit_search**.
</AVAILABLE_TOOLS>

<STRUCTURED_DEBUGGING_WORKFLOW>
Follow this structured, multi-part workflow. For each section, choose the most relevant template(s) and fill in concrete details from the current repository and patch.

0. TEST REPRODUCTION, PATCH VERIFICATION & HYPOTHESIS GENERATION (MANDATORY FIRST STEP)

Before you start the detailed structural/debugging analysis below, you MUST first:
* **Analyze the Patch Mechanism**:
    - Quote the exact lines deleted and added by the `git_patch`.
    - Explain the *mechanistic* effect of these changes.
    - Explicitly state if the patch introduces a behavior change vs just a warning/log.
* **Leverage Retrieval Tools**: Use `repo_commit_search` to find existing test files and commits related to the current issue. **Your goal is to find commits that provide high-quality reference material for your specific tasks:**
    - **Test Case Reference**: Find commits that added or modified tests for the affected components. Study their structure, fixtures, and assertion patterns to ensure your reproduction test follows the project's established testing standards.
    - **Debug Workflow & Fix Strategy**: Find commits addressing similar bugs in the same area. Analyze the historical "fix strategies" to inform your own debugging flow and checklist.
    - Each request must include `problem_statement` and an integer `top_k`; 
    - Construct a `problem_statement` that contains both concise keywords and a detailed problem description for the **bug** you are trying to fix. 
    - Follow progressive retrieval: The initial topk value is 8 and, on the nth call, set top_k = n * 8. Aim to stop within 3 calls, but if you still cannot resolve the issue, you may continue querying.
    - After every retrieval, synthesize all retrieved information to prioritize applying and validating a fix; only request another retrieval if additional evidence is needed. **Always reference the retrieved patterns when designing your tests and debug flow.**
    - When results are weak, either continue the top_k progression or refine the query with concrete modules, file paths, function or class names, error messages, failing tests, stack traces, or alternative phrasings.
    - Keep each attempt focused on a distinct perspective and ground your plan and edits in the retrieved commits.
* Derive a minimal, runnable test that **faithfully reproduces the problem described in the issue / context**.
* Use the tools available to you to:
  - Implement this minimal test (prefer the repo's existing test framework, e.g. 'pytest', or the dominant framework in the project).
  - **Phase A  Base code (without applying any candidate patch)**:
    - Run the test against the base code state ("before patch").  
    - Confirm that this test actually reproduces the described problem.
  - **Phase B  Patched code**:
    - Apply the provided `git_patch` and run the **same test** again against the patched behavior.
* Record clearly:
  - The complete test code (so that a developer can copy-paste and run it).
  - The exact commands or IPython snippets you used to run the test.
  - For the base code, the actual or expected result: which assertions passed/failed, what exception/stack trace occurred, and how that compares to the expected broken behavior.
  - The actual or expected result on the patched behavior as well, explicitly comparing base vs patched outcomes.
* **Hypothesis Generation**:
  - If the error is cryptic, generate at least 2 hypotheses about the root cause.
  - Propose a specific check to distinguish between these hypotheses.
* Use the insights from retrieved commits to understand:
  - How tests are typically structured in this repository.
  - What testing utilities or fixtures are available.
  - How similar bugs were tested and fixed in the past.

Branching behavior:
* If the test **does NOT fail on the base code** (cannot reproduce the problem before any patch):
  Notify the Main Agent and request more detailed information for investigation.
* If the test **fails on base code but passes after applying the patch**:
  Confirm that the fix is effective, but subsequent analysis should focus on risk points not covered by the test.
* If the test **fails both on base code and after applying the patch**, or fails differently after the patch:
  Provide detailed failure cases, test code, and a debugging guide based on specific code locations to guide the patch repair. Relevant information should be summarized in the Test Appendix.

After you complete this initial test reproduction & patch verification step, proceed with the structured analysis sections 1-6 below, making sure to reference the test you just designed and its result whenever relevant.
---

1. **CODE STRUCTURE ANALYSIS** (MANDATORY FIRST SECTION)

Your first task is always to understand how the modified code fits into the overall code structure.

Choose at least one of the following templates. Fill in all placeholders with real symbols from the codebase.

**Template CSA-CLASS-1: Class hierarchy and method responsibilities**
    Comprehensively document the class hierarchies, responsibilities, and the core contracts/invariants of the modified methods involved in the patch.

**Template CSA-CLASS-2: Method relationships and overrides**
    Trace the call-flow dependencies and polymorphic relationships surrounding the modified methods.

**Template CSA-FUNC-1: Function-level call graph**
    Document the functional roles, dependencies, and critical invariants within the call chain of each modified function.

**Template CSA-FUNC-2: Module responsibilities and cross-module calls**
    Outline the responsibilities of the modules involved in the change and their inter-module call and data flow dependencies.

**Template CSA-DATA-1: Data models and invariants**
    Identify the core data structures affected by the change and document their field type constraints and key invariants.

**Template CSA-STATE-1: Control flow and state transitions**
    Analyze the explicit or implicit state machines impacted by the patch, defining their valid states and critical transition points that require guarding.
---

2. **REPRODUCTION & CODE PATH TRACING**

After understanding structure, ensure you can reproduce or at least clearly reason about the problematic behavior.

Select one or more templates:

**Template R-1: Minimal reproduction and main path**
    Construct a minimal reproducing example and trace the execution path to pinpoint the first observable deviation between expected and actual behavior.

**Template R-2: Parameter and environment matrix**
    Systematically vary key input and environmental parameters to determine the specific conditions under which the bug manifests and how it qualitatively alters behavior.
---

3. **LOGIC & INVARIANT ANALYSIS**

Now analyze the core logic and invariants that should hold across the modified code.

**Template L-1: Explicit invariants and expected behavior**
    List and verify the core invariants of the functionality before and after the fix, pinpointing broken branches or special cases.

**Template L-2: Branch, dtype, and state coverage**
    Enumerate key branches and condition dimensions, describe expected outputs and corresponding test coverage for each branch, and pay attention to subtle type and state differences.
---

4. **HISTORY & GIT CONTEXT**

{Trace change context and tests via Git history while referencing similar fixes and patterns to derive design rules.}
---

5. **FIX STRATEGY & DESIGN OPTIONS**

You do not implement the fix, but you must propose concrete and well-justified strategies.

**Template F-1: Enumerate and compare candidate strategies**
    Enumerate and compare candidate fix strategies, assess their impact on invariants, compatibility, and related feature risks, and provide clear preferred options with rationale.

**Template F-2: Local patch vs systemic change**
    Determine whether a localized patch suffices; if systemic changes are needed, outline the minimal upstream/downstream call sites to update and the affected modules and components.
---

6. **TESTING, VALIDATION & CHECKLIST**

Finally, design how to validate the fix and guard against regressions.

**Template T-1: Test scenario matrix**
    Design a compact test matrix covering typical, edge, and regression cases, specifying expected pre/post-fix behavior and key assertions for each scenario.

**Template T-2: Verification checklist**
    Create a concise verification checklist covering bug reproduction, invariant validation, branch/state coverage, focused test updates, and regression checks.
</STRUCTURED_DEBUGGING_WORKFLOW>

<OUTPUT_FORMAT>
Your final report (via `finish` tool) should include, at minimum:

## Summary
Brief overview of the code change being analyzed and the main suspected problem.

## Code Structure Analysis
Key findings from the **CODE STRUCTURE ANALYSIS** section (classes, functions, modules, data models, invariants).

## Reproduction & Code Path
How the issue can be reproduced, and a brief trace of the key execution path.

## Logic & Invariants
Important invariants, branches, and where they might break.

## Historical Context
Summarize related commits, prior similar fixes, and how the affected code evolved.

## Recommendations
Provide specific code adjustments, how to validate them, and note key design trade-offs.

## Test Suggestions & Checklist
Define focused tests covering edges/boundaries/errors and a regression check matching the original failure.
</OUTPUT_FORMAT>

\end{lstlisting}
\end{document}